\documentclass[manuscript, screen, nonacm]{acmart}
\AtBeginDocument{%
  }

\usepackage{titlesec}
\titlespacing*{\section}{0pt}{*1.2}{*0.8}

\begin{document}

\title{Causal Inference Isn't Special: Why It's Just Another Prediction Problem}

\author{Carlos Fernández-Loría}
\email{imcarlos@ust.hk}
\orcid{0000-0003-4509-3768}
\affiliation{%
  \institution{Hong Kong University of Science and Technology}
  \city{Clear Water Bay}
  \state{New Territories}
  \country{Hong Kong}
}

\renewcommand{\shortauthors}{Fernández-Loría}



\maketitle


Causal inference is often seen as fundamentally different from predictive modeling. It introduces new terminology, new estimands, and what seems like a radically different goal: not predicting what \emph{will} happen, but what \emph{would} happen under alternative scenarios. This counterfactual framing---while essential---has given causal inference a reputation for being methodologically exotic and philosophically fraught.

However, from a modeling standpoint, the difference is smaller than it appears. In both cases, we start with labeled data drawn from a source domain---observations for which outcomes are known---and seek to generalize to a target domain where outcomes are unknown. In prediction, the target might be future behavior. In causal inference, it might be outcomes under a treatment condition that wasn't received. The goal in both cases is the same: estimate outcomes we do not observe. 

And the solution in both cases rests on the same foundation: \emph{assumptions} are required to justify generalization. Once this shared structure is made clear, causal inference no longer appears as a fundamentally different category of problem. It becomes what it is: a structured form of prediction.

This perspective does not downplay the complexity of causal inference. It \emph{is} special—a special case of prediction. And for that, we benefit from special tools: causal diagrams, potential outcomes, identification strategies~\citep{imbens2015causal,pearl2009causality}. These tools leverage the structure of the problem to clarify when generalization is justified. Indeed, it is the validity of this generalization---from observed treatment-outcome pairs to counterfactuals---that allows us to interpret statistical estimates causally. However, beneath the methodological differences lies the same core challenge as in predictive modeling: how do we move from observed outcomes in one domain to unobserved outcomes in another?

\section{Prediction and Generalization: The Usual Game}

In supervised learning, we build models by fitting them to labeled data—cases where both the inputs and the outcomes are observed. The ultimate goal, however, is not to perform well on this training data, but to generalize to new, unseen data where the outcomes are unknown. This might involve forecasting future events, but more generally, it means using patterns learned from past observations to estimate outcomes in situations we haven't seen before. Whether we're predicting next month's sales or whether a transaction is fraudulent, we're doing the same thing: using a model trained on known outcomes to infer the unknown.

The leap from the training domain to the deployment domain always relies on a crucial assumption: that the relationships learned from labeled cases will also hold in the unlabeled ones. This assumption is rarely guaranteed. The relationships between inputs and outcomes often shift between the training environment and the real-world setting where the model is applied. This is true whether the shift comes from time, geography, behavior, or other contextual changes. Even a model with perfect accuracy on the training data may fail completely in deployment if the conditions differ in subtle but important ways.

Despite this fragility, prediction remains useful because we often have good reason to believe that generalization is possible. In some cases, we assume that the relationship between inputs and outputs is stable enough for the model to transfer. In others, we take steps to correct for the differences between the source and target domains—through reweighting, domain adaptation, or incorporating domain knowledge~\citep{sugiyama2007covariate}. But the logic is the same: we learn patterns in one domain and apply them in another, guided by assumptions about generalizability.

\section{Enter Causal Inference: Same Game, Different Labels}

Causal inference is often framed around potential outcomes: the outcomes that would occur for an individual under treatment ($Y^1$) and under no treatment ($Y^0$)~\citep{rubin2005causal}. The individual-level effect is their difference, $Y^1 - Y^0$, and the goal is to estimate this effect for individuals or populations.

We formalize the problem with a treatment assignment variable $T \in \{0,1\}$, indicating whether each individual received the treatment ($T=1$) or not ($T=0$). For each unit, we observe the outcome under the condition they actually experienced: $Y^1$ if $T=1$, $Y^0$ if $T=0$.

A common claim is that the fundamental challenge of causal inference is that we only observe one of the two potential outcomes for each individual~\citep{holland1986statistics}. While literally true, this framing obscures the deeper modeling issue. In predictive modeling, we regularly predict outcomes we haven't seen. What makes causal inference hard is not the missing outcome---but that the observed outcome depends on treatment assignment, so the sample is potentially biased.

Consider estimating the average treatment effect (ATE). If we observed $Y^1$ for some individuals and $Y^0$ for others---both from the same target population---we could simply take the difference in means. This is what randomized experiments aim to approximate. While we still observe outcomes conditional on treatment assignment ($Y^1 \mid T=1$ and $Y^0 \mid T=0$), randomization increases our confidence that these conditional distributions approximate the marginal ones, $Y^1$ and $Y^0$.

But outcomes are always conditional on treatment assignment, so the data is always potentially biased. To infer $Y^1$ and $Y^0$, we must assume that patterns learned from $Y^1 \mid T=1$ and $Y^0 \mid T=0$ generalize to the full population. In randomized experiments, this generalization is more plausible because treatment assignment acts like random sampling from the full population. In observational studies, where treatment may depend on multiple factors, that sampling is often biased, so stronger assumptions and modeling tools---like causal diagrams, unconfoundedness assumptions, or instrumental variables---are needed to justify the leap.

The core challenge, then, is to justify generalizing from observed, treatment-conditional outcomes to unobserved potential outcomes. Causal inference has traditionally focused on estimating effects at the group or population level---such as the ATE---rather than predicting individual effects. But even these population-level estimates require assumptions to support generalization: we use data from the treated group to estimate what would have happened (on average) to those who were not treated, and vice versa.

Framed this way, causal inference becomes a structured instance of prediction under distribution shift: the shift comes from treatment assignment, but the task remains the same---using potentially biased observations to reason about outcomes we cannot observe directly. Whether we estimate the effect for a population or an individual, the logic is the same: we must predict the outcome under treatment and under no treatment. What enables a causal interpretation is our belief that the patterns learned from the treated and untreated groups can generalize to predict counterfactuals.

These structural parallels with predictive modeling become clearer when placed side by side. Table~\ref{tab:comparison} outlines the components of each framework, showing how causal inference fits naturally within the predictive paradigm---as a special case where training data is selectively sampled and the target variable is defined counterfactually.

\begin{table}[h]
  \centering
  \small
  \caption{Predictive Modeling vs. Causal Inference as Generalization Problems}
  \label{tab:comparison}
  \begin{tabular}{@{}p{2cm}p{5.8cm}p{5.8cm}@{}}
    \toprule
    \textbf{Component} & \textbf{Predictive Modeling} & \textbf{Causal Inference} \\
    \midrule
    Goal & Predict outcomes in a new context & Predict outcomes under different conditions \\
    Target Variable & $Y$ & $Y^1, Y^0$ \\
    Source Domain & Labeled data from a known context & Observed outcomes under one condition for those exposed to it (e.g., $Y^1 \mid T=1$) \\
    Target Domain & Unknown (unlabeled) outcomes in a different population or context & Unobserved outcomes under that condition for those not exposed to it (e.g., $Y^1 \mid T=0$) \\
    Challenge & Distribution shift (covariates, label drift) & Selection bias from treatment assignment \\
    Solution & Assumptions (stationarity, covariate shift) & Assumptions (unconfoundedness, overlap) \\
    Techniques & Reweighting, transfer learning, domain adaptation & IPW, covariate adjustment, matching \\
    \bottomrule
  \end{tabular}
\end{table}

\section{Causal Inference as Prediction Under Bias}

Whether in predictive modeling or causal inference, the core task is the same: using data where outcomes are observed to estimate outcomes in cases where they are not. In both cases, this leap is never justified by data alone---it rests on assumptions. Generalization is not something we \emph{observe}, but something we \emph{believe}, based on how we think the data-generating process works.

\begin{figure}
  \centering
  \includegraphics[width=0.7\linewidth]{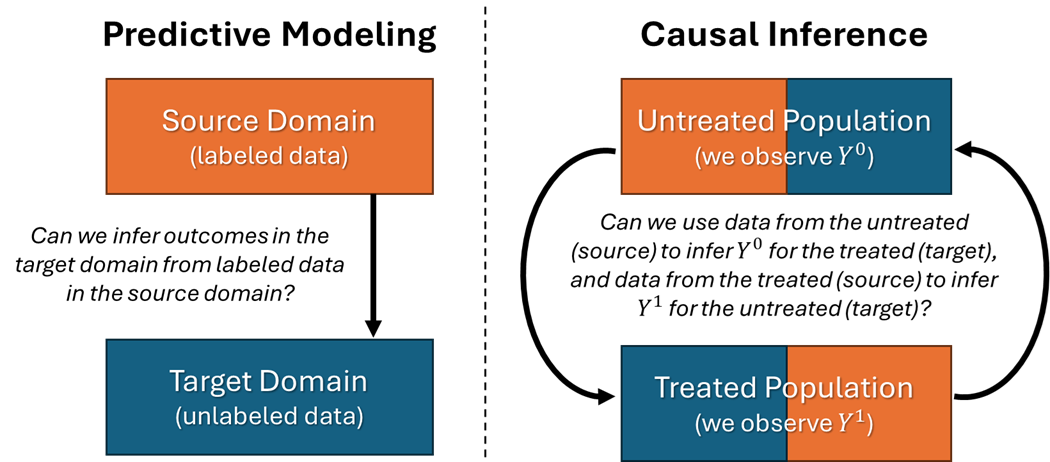}
  \caption{Both tasks involve inferring outcomes in an unseen domain using labeled data—and require assumptions to bridge the gap.}
 \Description{A side-by-side diagram comparing generalization in predictive modeling and causal inference.}
  \label{fig:generalization}
\end{figure}

Figure~\ref{fig:generalization} illustrates this shared structure. In predictive modeling, we use patterns from labeled data to estimate outcomes in unlabeled cases. In causal inference, we use outcomes from the treated to infer those of the untreated, and vice versa. In both cases, generalization relies on assumptions about how the domains are related.

In predictive modeling, these assumptions are often left implicit. We train a model on last year’s data and deploy it on this year’s, assuming customer behavior hasn’t changed. Or we transfer a model trained in one region to another, assuming stable relationships. These assumptions may go unstated, but they are central: if the conditions shift too much, the model fails.

Causal inference makes these assumptions explicit. To estimate what would happen under a different treatment condition, we must reason about how treatment assignment relates to outcomes. A common assumption is unconfoundedness: that, conditional on observed variables, treatment is as good as random~\citep{imbens2015causal}. Another is overlap: that everyone has a nonzero chance of receiving either treatment. These assumptions don’t make causal inference uniquely fragile---they simply make its requirements transparent.

Causal inference arguably offers a more honest accounting of what all generalization requires. We aim to infer outcomes in a domain where we lack observations, using data from a different, potentially biased domain. That's a familiar modeling challenge: learning from biased training data.

Much of causal methodology aims to correct this bias. Inverse probability weighting (IPW), for example, gives more weight to individuals underrepresented in their treatment group---much like reweighting in domain adaptation, where training examples are adjusted to better reflect the target domain. Covariate adjustment methods, such as regression or matching, aim to approximate what outcomes would look like if treatment were conditionally independent of potential outcomes. These techniques address selection bias to support valid generalization---just as machine learning methods correct for training set bias to improve performance in deployment settings.

\begin{figure}
  \centering
  \includegraphics[width=0.8\linewidth]{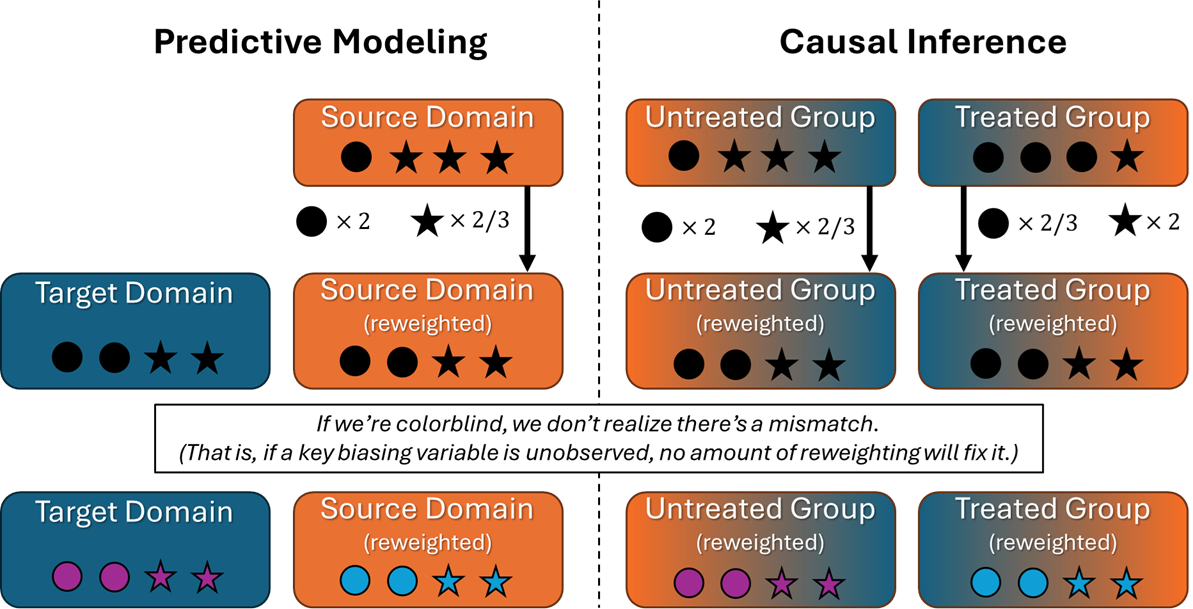}
  \caption{Correcting Sample Bias in Predictive and Causal Settings. Reweighting helps align source and target distributions—if the variables causing imbalance are observed.}
 \Description{A side-by-side diagram comparing techniques applied in predictive modeling and causal inference to correct for biased training data.}
  \label{fig:techniques}
\end{figure}

Figure~\ref{fig:techniques} illustrates this parallel. In predictive modeling, reweighting helps align the training data with the target distribution. In causal inference, reweighting or matching adjusts for differences in who receives treatment, so treated and untreated groups resemble each other in their covariate profiles. In both cases, the corrections rely on the same critical assumption: that we have access to variables that allow us to address the bias. If we're ``colorblind'' to an important source of bias---whether in domain adaptation or causal estimation---no method can fix the problem.

Ultimately, the logic is the same: we make assumptions, recognize bias, and apply modeling strategies to address it. The difference lies not in the challenge, but in the framing. Causal methodology is a toolkit---potential outcomes, causal diagrams, identification strategies---for reasoning about generalization. In that sense, it's not a fundamentally different task, but a structured instance of prediction under potential bias, where the assumptions for generalization are more clearly articulated. Our belief in the validity of that generalization is what allows us to interpret the estimates causally.

\section{What This Perspective Offers}

Viewing causal inference as prediction under bias offers several benefits:

\begin{itemize}
    \item \textbf{It sharpens conceptual understanding.} Focusing on the generalization challenge---rather than counterfactual language---reveals causal inference as the task of using selectively observed outcomes to predict outcomes under different conditions. Counterfactual reasoning remains central, but it need not obscure the predictive structure.
    \item \textbf{It encourages cross-pollination.} Techniques like reweighting, domain adaptation, and shift correction from machine learning offer ideas that can inform causal modeling. Conversely, causal reasoning can help predictive modelers clarify their assumptions and recognize when they may break down.
    \item \textbf{It supports better pedagogy.} For students trained in ML, causal inference often feels like entering a new and foreign world. But if we teach it as prediction with selectively sampled labels and structured assumptions, it becomes more intuitive. The challenge is familiar: making predictions where ground truth is missing.
    \item \textbf{It grounds causal inference in the practical realities of modeling.} It reminds us that assumptions are always required, and that generalization is always a leap---not a philosophical divide between causality and prediction, but a difference in inference targets under different forms of bias.
\end{itemize}

So yes---causal inference is about what would happen, not just what will. But the path to answering that question is one predictive modelers already know: build on what we've seen, understand what we haven’t, and make assumptions clear. The rest is just modeling.

\section*{Acknowledgments}

This article grew out of many conversations with Foster Provost about the fundamental differences---if any---between causal inference and prediction. My thinking was deeply shaped by these discussions, and this piece reflects the ideas that emerged from them. I am grateful to Foster for his insights, questions, and intellectual generosity throughout.

\bibliographystyle{ACM-Reference-Format}
\bibliography{references}
\end{document}